# Image Classification for Snow Detection to Improve Pedestrian Safety

**Ricardo de Deijn**
Minnesota State University, Mankato
**ricardodedeijn@gmail.com**

**Rajeev Bukralia**
Minnesota State University, Mankato
**rajeev.bukralia@mnsu.edu**

**ABSTRACT**

This study proposes a computer vision approach for snow detection on pavements aimed at reducing winter-related fall injuries, particularly for vulnerable demographics such as elderly and visually impaired individuals. Leveraging an ensemble of fine-tuned VGG-19 and ResNet-50 convolutional neural networks (CNNs), our approach achieves an accuracy of 72.7% and an F1 score of 71.8% on a custom dataset of 98 images created using a smartphone camera.

**INTRODUCTION**

According to Kakara, Moreland, Haddad, Shakya and Bergen (2021), there were at least 10,124 emergency department-treated fall injuries throughout the United States in 2015. The same happens every year to over 25,000 people in Sweden (Berggård, 2009). These injuries are most common in the age group 65 years and above. These high amounts of casualties could be significantly reduced through prevention and warning systems. Especially, since this age group is more likely to develop balance disabilities, compared to younger age groups, making a walk through the snow even harder (Burton, Reed and Chamberlain, 2011; Garcia, Pacheco, Villagrana, Aguilar and Aragon, 2018). While approaches for snow detection exist, an approach to detect snow from a handheld device for pedestrians on a sidewalk is yet to be found. A system that can detect snow using these devices might be of great help to improve pedestrian safety.. This paper proposes the idea of detecting snow in images of pavements through an ensemble of fine-tuned VGG-19 and ResNet-50 models, which are convolutional neural networks (CNNs).

In the following sections, we discuss our study's key aspects. We start with a review of related research, discuss our dataset and evaluation strategy, and explain the methodology. Then, we present our results, summarize conclusions, address limitations, and offer recommendations for future work.

**RELATED WORK**

Image classification, with its wide spectrum of applications, can improve pedestrian safety, especially for marginalized groups such as the elderly and visually impaired individuals. This is achieved by issuing, for example, auditory alerts as a safety measure when it detects potentially dangerous situations. (Mohamedon, Abd Rahman, Mohamad and Omran Khalifa, 2021). Recent advancements in computational power has resulted in a surge of applications using convolutional neural network (CNN) models trained on large datasets, tagged with specific targets for automatic classification (O'Shea and Nash, 2015).

Utilizing optical characteristics of snow in images, this study uses CNNs to recognize snow on pavements. Snow is made up of clustered ice crystals, giving it unique reflective properties due to how light interacts with these crystals. Coarse-grained snow absorbs light quickly due to longer travel distances in these crystals. It does need a bulk of snow, as a single ice crystal has a low absorption coefficient. Because of this, fine-grained snow tends to scatter light more often (Allgaier and Smith, 2022; Warren, 2019).

Previous research utilized various other wavelength sensor data, such as infrared, mmWave, and light wave, to detect snow and ice on pavements, to train CNN models with mixed successes (Abdalla, Iqbal and Shehata, 2017; Gailius and Jačėnas, 2007; Kim, Kim and Kim, 2022; Lee, Kang, Song and Hwang, 2020; Pooyoi, Borwarnginn, Haga and Kusakunniran, 2019). However, these approaches have *not yet* showcased the prospect of detecting snow by utilizing mobile devices to address the needs of disabled individuals to safely navigate pavements. To bridge this gap, this study employs VGG-19 and ResNet-50 CNN architectures on a custom dataset, utilizing prior work.

VGG-19 and ResNet-50, which are chosen for their pre-trained architectures on the ImageNet-1K dataset, excel in extracting high-level features from images (He, Zhang, Ren and Sun, 2015; Simonyan and Zisserman, 2014). With 19 and 50 weight layers, respectively, these models have demonstrated significant improvements in classification accuracy across diverse domains, and their architectures have functioned as inspiration to other CNN models used to detect road hazards (Lee et al., 2020). This makes them ideal candidates for this study.

**DATASET AND EVALUATION STRATEGY**

A self-produced dataset, created in winter 2024, consists of 98 images captured across various locations in Minnesota, the United States, using a Google Pixel 6a smartphone (3024 × 3024 pixels resolution). The images are evenly divided between *snow* and *snow-free* labels, taken at consistent locations and times to capture as much data diversity as possible within the limited amount of data. A separate test, containing images from distinct locations, ensures generalization evaluation of the model. The goal is an accurate classification of images as *snow* or *snow-free*. The evaluation strategy uses the F1 score and accuracy metrics on the test set predictions. F1 score combines precision (*true positives / predicted positives*) and recall (*true positives / actual positives*) metrics. The formula of the F1 score is:

$$F_1\ Score = 2 \times \frac{p \times r}{p + r}$$

(1)

The accuracy metric provides a straightforward measure of the model's overall correctness in predicting snow presence or absence. It does this by calculating the ratio of correctly classified images to the total number of images in the test set.

These two metrics are used to capture the general performance of the models' total prediction capabilities. However, to capture both false positives and negatives, the F1 score is used. This metric will show both the number of instances snow is incorrectly classified or ignored.

**METHODOLOGY**

**Implementation**

The VGG-19 and ResNet-50 model architectures are very good at extracting features. The implementation is done using PyTorch (Paszke, Gross, Massa, Bradbury, Lerer, Chanan, Lin, Killeen, Gimelshein, Desmaison, Antiga, Köpf, DeVito, Yang, Raison, Chilamkurthy, Tejani, Steiner, Bai, Fang and Chintala, 2019). This work uses transfer learning on the pre-trained models, which is an approach that handles small data set sizes well, due to the reuse of pre-trained feature-extract configurations (Mohamedon et al., 2021). After transfer learning, the models are ensembled through a weighted averaging method, which combines the best predictions for both models by adding weights to the best-performing model. After ensembling, the function returns one average conclusive prediction (An, Ding, Yang, Au and Ang, 2020; He et al., 2015).

**Experiments**

Before model fine-tuning, the images are resized to a resolution of 128x128 pixels and normalized to the range [0,1]. This is followed by further normalization based on the mean [0.485, 0.456, 0.406] and standard deviation [0.229, 0.224, 0.225] of the models' training dataset, ImageNet-1K (Torch Contributors, 2017). Model training uses the Adam optimizer to optimize the classification weights in the Fully Connected layers (Kingma and Ba, 2014), using initial learning rates ([0.0001, 0.001, 0.01, 0.1]) and epoch counts ([15, 20, 25]). The other Adam parameters are not tuned and thus use the PyTorch defaults (Torch Contributors, 2023). The original architecture weights of the models remain unchanged. However, the classification layer is replaced and optimized in both models to allow predictions of *snow* or *snow-free* as an output classification. Among the tested learning rates, 0.0001 showed the most promising results across all epoch counts.

**Training and Inference**

For model fine-tuning, a batch size of 4 is selected to balance convergence with computational efficiency. This study uses an NVIDIA GeForce GTX 1650 GPU with 4GB of memory, providing sufficient computational power. The dataset is divided into 80% training and 20% validation subsets, each containing paired *snow* and *snow-free* images from identical locations to facilitate quicker learning. Additionally, a separate test set containing 22 images from distinct locations is created, with *snow* and *snow-free* images not paired up**.** This enables the model to showcase its generalization capabilities under diverse conditions.

**RESULTS AND ANALYSIS**

While finetuning the two convolutional neural network architectures for 25 epochs, the trained models save every 5 epochs. After training was completed, every saved model was tested using the test set, and its F1 score and accuracy score were determined. Epochs 5 and 10 showed insignificant F1 scores and accuracy results. Epochs 15, 20, and 25 however showed a minor stable improvement for both the F1 scores and accuracy evaluation metrics on the individual models, as shown in Table 1.

| Epochs | Loss | Acc | Val Loss | Val Acc | F1 Score on Test Set | Accuracy on Test Set | FP/P | FN/N |
|---|---|---|---|---|---|---|---|---|
| 15 | 0.351 | 1.990 | 0.591 | 1.778 | Ensemble: 71.8% | Ensemble: 72.7% | 45.5% | 27.3% |
|  |  |  |  |  | VGG-19: 033.3% | VGG-19: 50.0% | 100% | 0% |
|  |  |  |  |  | ResNet-50: 50.9% | ResNet-50: 54.5% | 18.2% | 72.7% |
| 20 | 0.286 | 1.987 | 0.538 | 1.778 | Ensemble: 40.0% | Ensemble: 50.0% | 90.9% | 9.1% |
|  |  |  |  |  | VGG-19: 31.3% | VGG-19: 45.5% | 9.1% | 100% |
|  |  |  |  |  | ResNet-50: 31.3% | ResNet-50: 45.5% | 100% | 9.1% |
| 25 | 0.243 | 1.987 | 0.501 | 1.833 | Ensemble: 45.5% | Ensemble: 45.5% | 54.5% | 54.5% |
|  |  |  |  |  | VGG-19: 40.0% | VGG-19: 50.0% | 9.1% | 90.9% |
|  |  |  |  |  | ResNet-50: 50.9% | ResNet-50: 54.5% | 72.7% | 18.2% |

**Table 1. Summary of training results (with learning rate 0.0001)**

However, the best achieved results were achieved using the ensemble of VGG19 and ResNet50 after 15 epochs. This system was able to achieve an accuracy of 72.7% and an F1 score of 71.8% on the unseen test dataset. Table 1 also shows that epoch 15 showed the best ratio of false positives (FP) compared to all positives (P) and false negatives (FN) compared to all negatives (N). This is also shown in figure 1, where examples of misclassified images are represented in **bold**. This figure shows examples of false negatives predictions, as well as false positives.

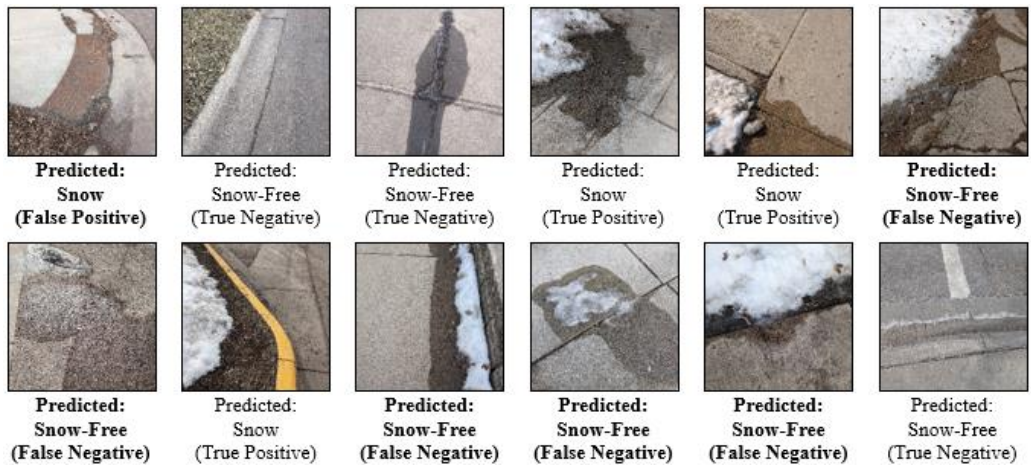

**Figure 1. Example of several test image classifications. The texts in bold represent misclassified images.**

False negatives appear less frequent in epoch 15. However, false negatives form a higher risk for pedestrians when actively used. Figure 1 shows a couple of situations where false negatives appear, as well as false positives. Insufficient variation in snowy conditions in the training data may be the primary cause for this phenomenon to happen, and got amplified by the number of variating images in the test dataset.

Besides this, the ResNet-50 and VGG-19 models are not optimized on data showing subtle changes in light wave reflections. Due to this, the model had limited ability to grasp subtle details, such as the way light reflects off snow or the appearance of pavement in snowy situations. Its performance could be enhanced by training it anew on a dataset specifically designed to emphasize reflections.

**CONCLUSION**

In conclusion, this study presents an approach for snow detection on sidewalks and pavements from a custom dataset containing 98 smartphone images. It uses an ensemble of fine-tuned VGG-19 and ResNet-50, which are high-level feature-extracting models with a track record in hazard detection. Through experimentation and evaluation, the study shows that the proposed methodology achieved promising results, with the ensemble model returning an accuracy of 72.7% and an F1 score of 71.8% on the unseen test dataset.

Misclassification, especially in snow-related images, stems from both the limited training data on *snow* and *snow-free* pavements as well as an absence of light wave reflection context in the pre-trained VGG-19 and ResNet-50 models. Allgaier and Smith (2022) highlight the inherent difficulty in predicting based on light wave reflections. This necessitates more data and optimized model architectures specialized for such reflections.

Despite these hurdles, the study highlights the potential of image classification systems in reducing winter-related risks, particularly for vulnerable demographics like the elderly and visually impaired.

**LIMITATIONS AND FUTURE WORK**

Looking ahead, there is potential for improvement in the dataset used for snow detection. Models like ResNet-50 and VGG-19 were not initially designed to identify reflections of light waves on snow. To address this, it is necessary to train these or other models from the ground up with a dataset that focuses on the reflective properties of light on snow. Another addition to the dataset is to concatenate the temperature at the time the image is taken to the CNN feature maps just before the dense layers. These changes would improve the model's adaptability to different scenarios and lead to significant advancements in real-time snow detection systems, contributing to pedestrian safety during the winter. Additional improvements involve significantly expanding the dataset with images taken by different types of cameras, in varied locations, under diverse lighting conditions, and on various types of pavements. Further enhancements might include adjusting model parameters such as batch size, weight decay, and adding different types of noise to the data. These steps can help prevent overfitting and boost the overall effectiveness of the models.

A last improvement would involve using more false negative focused evaluation metrics, such as recall or F2 score. This research focusses on classifying snow and no-snow, however more attention could be given to when a model misclassifies snow as no-snow instead of using an equal weight between the two classes.

**ACKNOWLEDGMENTS**

I would like to acknowledge that this research/project was conducted without external funding or grants.